\documentclass{article}

\usepackage[main, final]{neurips_2026}

\usepackage[utf8]{inputenc} 
\usepackage[T1]{fontenc}    
\usepackage{hyperref}       
\usepackage{url}            
\usepackage{booktabs}       
\usepackage{amsfonts}       
\usepackage{nicefrac}       
\usepackage{microtype}      
\usepackage{xcolor}         
\usepackage{graphicx}
\usepackage{amsmath}
\usepackage{multirow}
\usepackage{bm}
\usepackage{adjustbox}
\usepackage{wrapfig}
\usepackage{capt-of}

\title{\system: IO-aware Serving for Linear Attention}

\author{%
  Longwei Zou \\
  Department of Computer Science\\
  Yale University\\
  New Haven, CT 06511 \\
  \texttt{longwei.zou@yale.edu} \\
  \And
  Lin Zhong \\
  Department of Computer Science\\
  Yale University\\
  New Haven, CT 06511 \\
  \texttt{lin.zhong@yale.edu} \\
}

\usepackage{color,soul}
\usepackage{xspace}

\definecolor{lightgray}{gray}{0.9}
\definecolor{gray}{gray}{0.85}
\definecolor{lightblue}{rgb}{0.9,0.9,1}
\definecolor{blue_bg}{rgb}{0.7,0.85,1}
\definecolor{lightyellow}{rgb}{1,1,0.8}
\definecolor{lightpurple}{rgb}{1,0.85,1}
\definecolor{red}{rgb}{1,0,0}
\definecolor{darkgreen}{rgb}{0.4,0.7,0.3}
\definecolor{darkblue}{rgb}{0.2,0.7,0.9}

0in \topmargin -.3in \advance \topmargin-\baselineskip \headsep .5in
\makeatletter

\renewcommand\section{\@startsection {section}{1}{\z@}%
  {-2ex\@plus -1ex \@minus -.2ex}%
  {0.3ex \@plus .2ex}%
  {\normalfont\large\bfseries}}
\renewcommand\subsection{\@startsection{subsection}{2}{\z@}%
  {-1.5ex\@plus -1ex \@minus -.2ex}%
  {0.5ex \@plus .2ex}%
  {\normalfont\normalsize\bfseries}}
 
 \renewcommand\subsubsection{\@startsection{subsubsection}{3}{\z@}%
  {-1ex\@plus -1ex \@minus -.2ex}%
  {0.3ex \@plus .2ex}%
  {\normalfont\normalsize\bfseries}}
\renewcommand{\paragraph}{%
  \@startsection{paragraph}{4}%
  {\z@}{0.5ex \@plus .2ex \@minus .2ex}{-1em}%
  {\normalfont\normalsize\bfseries}%
}
\makeatother

\let\oldhat\hat
\renewcommand{\hat}[1]{\oldhat{\mathbf{#1}}}

\definecolor{lst_py_kw}{rgb}{0.10,0.20,0.70}    
\definecolor{lst_py_str}{rgb}{0.70,0.10,0.10}   
\definecolor{lst_py_cmt}{rgb}{0.00,0.45,0.00}   

\definecolor{lst_rs_kw}{rgb}{0.70,0.35,0.00}    
\definecolor{lst_rs_str}{rgb}{0.65,0.15,0.15}   
\definecolor{lst_rs_cmt}{rgb}{0.00,0.45,0.00}   
\definecolor{lst_frame}{gray}{0.80}

\newcommand{\system}[0]{\texttt{KVBuffer}\xspace}

\begin{document}

\maketitle

\begin{abstract}
    
Linear attention has recently gained significant attention for long-context inference due to its constant decoding cost with respect to context length. 
However, existing serving systems typically serve linear attention by recurrently computing and updating a large linear attention state in every decoding step. Since the state is much larger than the per-token key and value, recurrent decoding incurs substantial memory access and becomes inefficient for serving linear attention. 
In this paper, we propose \system, an IO-aware serving mechanism for linear attention. By buffering recent keys and values, \system enables serving systems to compute linear attention outputs in more flexible and memory-efficient ways. For decoding, \system enables chunkwise computation, which reduces average memory access and decoding latency by deferring state updates and applying them in batch. For speculative decoding, \system verifies draft tokens in parallel and avoids storing temporary states. For short contexts, \system computes attention outputs directly from buffered keys and values, without creating or updating the linear attention state. 
We implement \system in SGLang for Qwen3.5 and Nemotron-3 series models. Our evaluations show that \system can speedup batch decoding by up to $1.47 \times$ and increase the maximum number of serving requests by $5 \times$ for speculative decoding when verifying four draft tokens.

\end{abstract}

\section{Introduction}
In recent years, long-context workloads have become increasingly prevalent in the applications of Large Language Models (LLMs), particularly agentic applications. To address the growing demand for efficient long-context processing, linear attention has attracted growing interest due to its constant decoding cost and bounded GPU memory footprint with respect to context length. With techniques such as gating\cite{GLA}, decaying mechanisms\cite{Mamba} and delta rule\cite{DeltaNet, GatedDeltaNet}, linear attention-based LLMs have substantially narrowed the quality gap with those using softmax-based attention. 
As a result, many recent LLMs\cite{Qwen3Next, KimiLinear, MiniMaxM1} employ hybrid architectures that interleave linear and softmax attentions to balance model quality and inference efficiency, achieving improved trade-offs over standard Transformer designs.

Existing serving systems\cite{PagedAttention,SGLang}, unfortunately, are very inefficient for linear attention-based models, because they compute the linear attention state recurrently (\S\ref{sec:background_serving_la}).
Typically, a serving system maintains one linear attention state for each request. 
After prefilling the prompt into linear attention state, the serving system updates the state with the newly generated key and value in each decoding step, then produces the attention layer output by querying the updated state.
The inefficiency stems from the large size of the linear attention state and how it is updated. The linear attention state is usually larger than the per-token key and value (KV) by two orders of magnitude, e.g., 2MB in Qwen3-Next Gated DeltaNet Layer\cite{Qwen3Next}.
Existing serving systems read and write the linear attention state in every decoding step, which consumes a substantial portion of the memory bandwidth and is sequential in nature.
The problem is even more pronounced in two common cases. 
In speculative decoding, the serving system must recurrently compute a temporary linear attention state for each draft token and store these temporary states until draft tokens are verified, further multiplying memory and memory access by the number of draft tokens. For example, verifying $4$ draft tokens occupies an additional 294.75MB of memory per-request in the Qwen3-Next model, imposing a substantial burden on GPU memory. 
For short contexts, the linear attention state can use more memory than the keys and values (KVs) of all tokens. In this case, computing attention output directly from all KVs can be more memory-efficient than recurrently maintaining the linear attention state.

In this work, we present \system, an IO-aware serving mechanism for linear attention. 
The key idea is to buffer the KVs of consecutive tokens and update the linear attention state in batch, in parallel, instead of recurrently.
Although this incurs additional memory use, especially bandwidth use, by buffered KVs,
it is more than offset by the savings from updating the linear attention state less frequently. As shown in \S\ref{sec:kvbuffer_chunkwise_decode}, by balancing the increased memory reads to buffered keys and values against the reduced memory writes from less frequent state updates, \system minimizes the average memory access per decoding step with a buffer size of $2\sqrt{d}$, where $d$ is the hidden dimension.
This memory efficiency, however, comes with a latency trade-off. When the buffer is not full, \system avoids updating the linear attention state and thus reduces per-token decoding latency. Once the buffer is full, \system must flush the buffered KVs and update the linear attention state, introducing latency for that decoding step. Since the state is updated with buffered KVs in parallel on GPU, the latency remains modest and does not scale linearly with the buffer size (See \S\ref{sec:kvbuffer_chunkwise_decode}).
For speculative decoding, instead of recurrently computing and creating a temporary linear attention state for each draft token, \system supports parallel verification of multiple draft tokens by buffering their corresponding keys and values, followed by a single update to the linear attention state using only the accepted tokens. 
For contexts with $d$ or fewer tokens, \system buffers all KVs and directly computes the attention layer output from KVs, \emph{without} ever creating or updating a linear attention state in memory. 
Note in this case, the buffer size can be up to $d$, instead of $2\sqrt{d}$.

We implement \system in SGLang for Qwen3.5 and Nemotron-3 series models. Qwen3.5 and Nemotron-3 both adopt a hybrid architecture that incorporates Gated DeltaNet and Mamba-2 as their linear attention modules, respectively. Experimental results show that \system speedup batch decoding by up to $1.47 \times$. For speculative decoding, \system reduces verification latency and increases the maximum number of serving requests as the number of draft tokens increases, improving end-to-end serving throughput by up to $1.9\times$. Finally, we demonstrate that KV-only decoding is more efficient than both recurrent and chunkwise decoding for short-context requests.
We also implement \system in Pie, a programmable serving system~\cite{Pie}, that allows users to explicitly manage linear attention states, as detailed in \ref{sec:implementation}.

\section{Background}

\subsection{Linear Attention and Its Computation Forms}
\label{sec:background_la}

Let the sequence length be $L$ and the hidden dimension be $d$. Standard softmax attention computes the attention output as:

\begin{align}
    \mathbf{Q}, \mathbf{K}, \mathbf{V} &= \mathbf{X}\mathbf{W}_Q, \mathbf{X}\mathbf{W}_K, \mathbf{X}\mathbf{W}_V \notag \\
    \mathbf{O} &= \text{Softmax}((\mathbf{Q}\mathbf{K}^T) \odot \mathbf{M})\mathbf{V} \notag 
\end{align}

where $\mathbf{X} \in \mathbb{R}^{L \times d}$ is the input, $\mathbf{W}_Q, \mathbf{W}_K, \mathbf{W}_V \in \mathbb{R}^{d \times d}$ are learnable parameters, and $\mathbf{M}$ denotes the causal mask defined as $\mathbf{M}_{ij}=1$ when $j \leq i$, otherwise $-\infty$. During inference, softmax attention needs to store keys and values of all previous tokens, gradually increasing the memory occupation and access with regard to the context length.

Linear attention\cite{LA} removes the softmax operation and computes attention output as follows. For simplicity, we omit the normalization and query/key feature maps.

\begin{align}
    \mathbf{Q}, \mathbf{K}, \mathbf{V} &= \mathbf{X}\mathbf{W}_Q, \mathbf{X}\mathbf{W}_K, \mathbf{X}\mathbf{W}_V \notag \\
    \mathbf{O} &= ((\mathbf{Q}\mathbf{K}^T) \odot \mathbf{M})\mathbf{V} \label{eq:parallel_la}
\end{align}

We refer to Eq. \ref{eq:parallel_la} as the \emph{parallel form} of linear attention, because the attention outputs of all tokens can be computed simultaneously. For a single decoding step $t$, the same computation can be written as: 

\begin{align}
    \bm{q}_t, \bm{k}_t, \bm{v}_t &= \bm{x}_t \mathbf{W}_Q, \bm{x}_t \mathbf{W}_K, \bm{x}_t \mathbf{W}_V \notag \\
    \bm{o}_t &= \sum_{i=0}^{t}{\bm{q}_t \bm{k}_i^T \bm{v}_i} \label{eq:parallel_la_decoding}
\end{align}

where $\bm{x}_t, \bm{o}_t, \bm{q}_t, \bm{k}_t, \bm{v}_t \in \mathbb{R}^{1 \times d}$ denote the input, output, query, key, and value vectors of the $t$-th token. For convenience, we also refer Eq. \ref{eq:parallel_la_decoding} as parallel form.

Eq. \ref{eq:parallel_la_decoding} can also be computed \emph{recurrently} by maintaining a fixed-size state:

\begin{align}
    \mathbf{S}_t &= \mathbf{S}_{t-1} + \bm{k}_t^T \bm{v}_t \notag \\
    \bm{o}_t &= \bm{q}_t \mathbf{S}_t \label{eq:recurrent_la_decoding}
\end{align}

where $\mathbf{S}_t \in \mathbb{R}^{d \times d}$ is the linear attention state at step $t$. Compared with the parallel form, the \emph{recurrent form} avoids storing and accessing all previous keys and values. As a result, it has constant memory access and decoding latency, which makes it attractive for long-context inference.

By expanding the recurrent form over $m$ steps, we can derive the \emph{chunkwise form} \cite{GLA} as follows: 

\begin{align}
    \bm{o}_j &= \bm{q}_j \mathbf{S}_{t-m} + \sum_{i=t-m+1}^{j}{\bm{q}_j \bm{k}_i^T \bm{v}_i} , \quad \text{for } j = t-m+1,\ldots,t \label{eq:chunkwise_la_decoding} \\
    \mathbf{S}_t &= \mathbf{S}_{t-m} + \sum_{i=t-m+1}^{t}{\bm{k}_i^T \bm{v}_i} \label{eq:chunkwise_la_state_update}
\end{align}
The chunkwise form computes the attention output $\bm{o}_j$ for token $j$ using both the linear attention state of step $t-m$ and the keys and values of tokens $t-m+1$ through $j$, as shown in Eq.~\ref{eq:chunkwise_la_decoding}. After every $m$ decoding steps, it updates the state using the keys and values of the most recent $m$ tokens, as shown in Eq.~\ref{eq:chunkwise_la_state_update}. Therefore, it allows state updates to be deferred and applied in batches, which is useful for reducing average memory access during decoding.
The parallel and recurrent forms can be considered special cases of this chunkwise form. 

Several variants\cite{GLA, GatedDeltaNet, Mamba2, KimiLinear} of linear attention have been developed to improve its long-context retrieval performance. The three computation forms discussed above are also applicable to these variants. In this work, we use Gated Delta Networks (GDN)\cite{GatedDeltaNet} and Mamba-2\cite{Mamba2} for evaluation, since they have been widely adopted in recent hybrid models\cite{Qwen3Next, KimiLinear}. We provide more details on GDN and Mamba-2 in Appendix~\ref{sec:gated_delta_net} and \ref{sec:mamba_2}, repsectively.

\subsection{Serving with Linear Attention}
\label{sec:background_serving_la}

When serving linear attention-based models, existing serving systems, such as vLLM and SGLang, allocate and maintain a linear attention state in the memory pool for each request. After prefilling the prompt into the linear attention state, they recurrently update the state and compute the attention output by Eq. \ref{eq:recurrent_la_decoding}, i.e., using the recurrent form. Moreover, in speculative decoding, the serving system recurrently computes a temporary linear attention state for each draft token and stores these temporary states in the state pool, which incurs additional consumption of $N$ state slots, where $N$ is the number of draft tokens.

\begin{table}[htbp]
    \centering
    \caption{Memory storage and average per-token memory access during decoding for different forms of linear attention computation. $L$ denotes the context length, $d$ the hidden dimension, and $m$ the chunk size. Memory access is measured in bytes, assuming that the linear attention state is stored in FP32 and $\bm{q}, \bm{k}, \bm{v}, \bm{o}$ are stored in FP16. For the recurrent form, the state update and attention output computation are fused into a single kernel.}
    \begin{tabular}{c|c|c|c}
        \toprule
        \multirow{2}{*}{Computation Forms} & \multirow{2}{*}{Memory Storage} & \multicolumn{2}{c}{Memory Access} \\
        \cline{3-4}
        & & Read & Write \\
        \hline
        Parallel & $4Ld$ & $4Ld + 2d$ & $6d$ \\
        \hline
        Recurrent & $4d^2$ & $4d^2 + 6d$ & $4d^2 + 2d$ \\
        \hline
        Chunkwise & $4d^2 + 4md$ & $4(1+1/m) d^2 + 2(m+4)d$ & $4d^2 / m + 6d$ \\
        \bottomrule
    \end{tabular}
    \label{tab:la_mem_profile}
\end{table}

As shown in Table \ref{tab:la_mem_profile}, the three computation forms of linear attention have different trade-offs among memory storage, memory read and memory write during decoding.
The parallel form stores the keys and values of all previous tokens, so both its storage and memory reads grow with the context length $L$. However, it avoids maintaining the linear attention state, making it more efficient when the context length $L < d$.
The recurrent form maintains a fixed-size state, giving constant storage and memory access with respect to the context length. This property makes it suitable for long-context decoding and thus is adopted in existing serving systems during decoding.
However, recurrent decoding requires reading and writing the full state in every decoding step, which is expensive because the state is $d$ times larger than a per-token KV.
Compared to recurrent form, chunkwise form introduces additional memory access from reading the buffered keys and values of recent tokens for each decoding step, but it amortizes state updates across multiple tokens and thus reduces average memory access.

Considering memory access, recurrent form is not always the most efficient choice for serving linear attention. In \S \ref{sec:kvbuffer}, we introduce \system mechanism. By buffering keys and values, \system allows us to flexibly select the most efficient computation form for different decoding scenarios.

\section{\system}
\label{sec:kvbuffer}

We now describe the design of \system, an IO-aware serving mechanism that buffers recently generated keys and values to enable more flexible computation forms for linear attention decoding.
We first overview \system and its paged memory management (\S\ref{sec:kvbuffer_design}).
We then show how \system supports chunkwise decoding, which reduces average memory access and decoding latency (\S\ref{sec:kvbuffer_chunkwise_decode}).
Next, we describe how \system enables parallel verification of draft tokens in speculative decoding (\S\ref{sec:kvbuffer_spec_decoding}).
Moreover, we demonstrate that \system also supports decoding in parallel form, which computes attention output only with KVs and is more efficient than both recurrent and chunkwise forms for short contexts (\S\ref{sec:kvbuffer_parallel_decode}). 
Finally, we show details of implementation in SGLang and Pie serving systems.

\begin{figure}[htbp]
    \centering
    \includegraphics[width=0.75\linewidth]{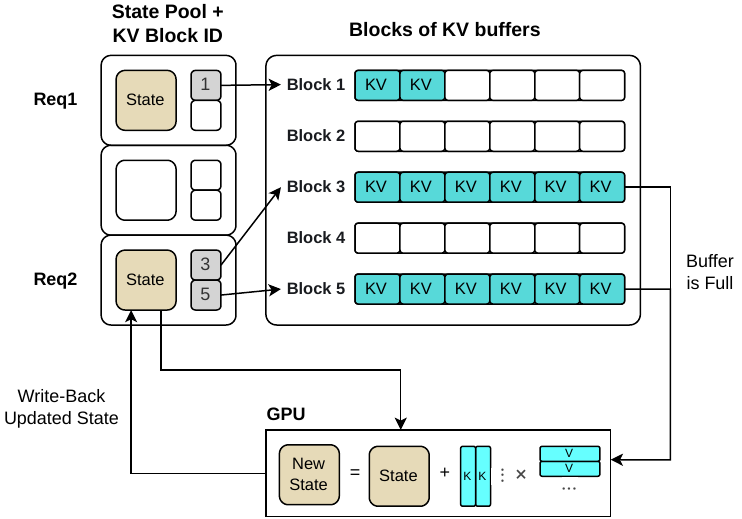}
    \caption{\system Design. We partition the memory for KV buffers into blocks, each of which can store $6$ KVs. 
    Each request has two blocks. During decoding, the serving system loads state along with buffered KVs to compute attention output. When the buffer is full, the state is updated with all buffered KVs on GPU and the updated state will be written back to the state slot.}
    \label{fig:kvbuffer_rep}
\end{figure}

\subsection{Design Overview}
\label{sec:kvbuffer_design}

As shown in Figure \ref{fig:kvbuffer_rep}, in addition to the linear attention state, allocated from a pool like existing serving systems, \system introduces a buffer of KVs for each request. To accommodate dynamically growing KVs, \system draws inspiration from paged attention\cite{PagedAttention} and allocates the KV buffer for a request from a pool of memory shared system-wide. The basic unit of allocation is a block whose capacity is configured by the user.
This design allows the KV buffer of a request to be dynamically sized and avoids memory fragmentation, thereby efficiently supporting linear attention decoding in chunkwise and parallel forms.

\subsection{Chunkwise Decoding with \system}
\label{sec:kvbuffer_chunkwise_decode}

By using the linear attention state together with the buffered KVs, \system supports chunkwise decoding, which is more memory efficient than the recurrent form.
In the prefilling stage, the serving system computes the prompt into the linear attention state. During decoding, the system loads the state along with buffered KVs to compute the linear attention output using Eq. \ref{eq:chunkwise_la_decoding}. It places newly generated KVs in the KV buffer. 
When the KV buffer is full, \system updates the linear attention state with all buffered KVs in batch, according to Eq. \ref{eq:chunkwise_la_state_update}.

Ideally, when both the recurrent and chunkwise forms are memory-bound, their decoding latency is proportional to the number of memory accesses. Therefore, we can estimate the speedup by the memory access ratio between recurrent and chunkwise forms.
Following Table \ref{tab:la_mem_profile}, we assume that the linear attention state is stored in FP32, and $\bm{q}, \bm{k}, \bm{v}, \bm{o}$ vectors are stored in FP16. Without loss of generality, we assume there is a single attention head. The speedup is shown as follows:

\begin{align}
    \text{Speedup}_{\text{chunkwise\_decoding}}(m) &\approx \frac{4(d+1)}{2d + \frac{4d}{m} + m + 7} \label{eq:kvbuffer_decoding_speedup}
\end{align}

where $m$ is the chunk size and $d$ is the hidden dimension. The ideal speedup is maximized at $m^*=2\sqrt{d}$, which corresponds to the optimal KV buffer size for chunkwise decoding with \system. When the memory bandwidth of GPU is high enough, the chunwise decoding could be compute-bound. Therefore, the buffer size $m$ should be selected based on preliminary sweeping experiments. We evaluate different settings of GPUs and details the buffer size $m$ selection in \ref{sec:expr_chunkwise_decoding}. 

On the other hand, while \system reduces per-token decoding latency, it must flush the buffered KVs when the buffer is full, which incurs additional state update latency for that step.
However, since the state update over all buffered KVs is performed in parallel on the GPU, the latency remains modest and does not scale linearly with the buffer size. 
Following the same assumptions in Table \ref{tab:la_mem_profile}, the arithmetic intensity of the state update in Eq. \ref{eq:chunkwise_la_state_update} is ${m}/{(4 + 2m/d)}$. When $m$ is not very large, e.g., $m=2\sqrt{d}$, the state update is still memory-bound. Consequently, its latency is approximately proportional to its memory access, $8d^2 + 4md$, where the buffer size dependent term $4md$ is relatively small compared with the state access term $8d^2$ for the optimal buffer size.

\subsection{Parallel Verification for Speculative Decoding}
\label{sec:kvbuffer_spec_decoding}

Speculative decoding is an important technique to accelerate LLM decoding by verifying multiple draft tokens in parallel~\cite{SpecDecoding, EAGLE}.
However, existing serving systems for linear attention incur significant memory overhead by maintaining and recurrently computing a temporary linear attention state for each draft token, until accepted tokens are determined.

\begin{figure}
    \centering
    \includegraphics[width=0.85\linewidth]{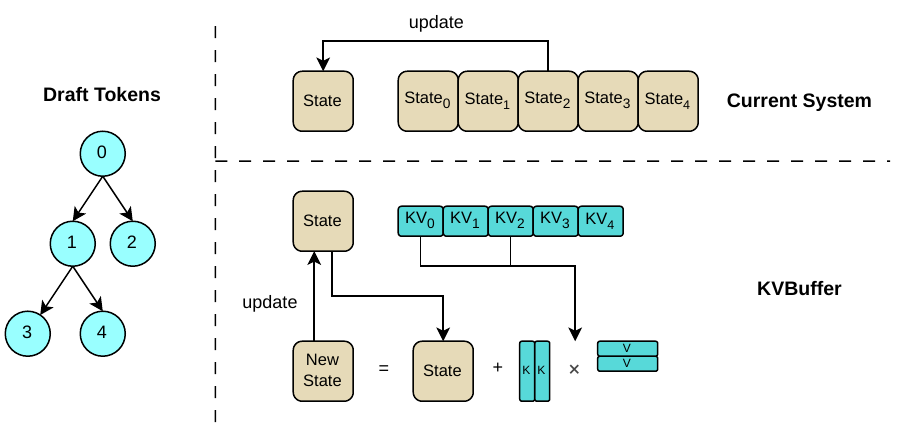}
    \caption{Speculative Decoding with \system. Existing serving systems allocate a temporary linear attention state for each draft token and use the recurrent form for speculative decoding verification. After determining that the accepted draft tokens are $0$ and $2$, the state of this request is replaced by the temporary $\text{state}_2$. In contrast, \system buffers the KV for each draft tokens and updates the state with KVs of only accepted tokens, reducing overall memory use.}
    \label{fig:parallel_verification}
\end{figure}

\system enables memory-efficient parallel verification of draft tokens. During verification, the serving system can buffer KVs of draft tokens, and  compute their attention outputs using the chunkwise form in Eq. \ref{eq:chunkwise_la_spec_verify}. After having determined accepted tokens, the serving system updates the linear attention state with buffered KVs of only accepted tokens, as in Eq. \ref{eq:la_spec_state_update} and Figure \ref{fig:parallel_verification}, reducing overall memory use.

\begin{align}
    \mathbf{O} &= \mathbf{Q} \mathbf{S}_{t-j} + ((\mathbf{Q} \mathbf{K}^T) \odot \mathbf{M}) \mathbf{V} \label{eq:chunkwise_la_spec_verify} \\
    \mathbf{S}_t &= \mathbf{S}_{t-j} + \mathbf{K}_{acc}^T \mathbf{V}_{acc} \label{eq:la_spec_state_update}
\end{align}

Without loss of generality, we suppose that $m$ draft tokens are sequential and all accepted. The arithmetic intensities of attention output computation and state update in Eq. \ref{eq:chunkwise_la_spec_verify} and Eq. \ref{eq:la_spec_state_update} are $\frac{2md^2 + 2m(m+1)d}{4d^2 + 12md}$ and $\frac{m}{4+2m/d}$, respectively, both of which indicate memory-bound operations when $m$ is relatively small. Therefore, the expected speedup can be approximated by the ratio of memory access, given by Eq. \ref{eq:spec_verify_speedup}, which approaches $\frac{m+1}{3}$ when $d>>m$. 

\begin{align}
    \text{Speedup}_{\text{parallel\_verify}} (m) \approx \frac{(m+1)d+2m}{3d+4m} \label{eq:spec_verify_speedup}
\end{align}

Note that for the recurrent form, we estimate the memory access by assuming that the verification steps are fused into a single kernel and we only read the initial state $S_{t-j}$ once because draft tokens are sequential. For $m=2$, speculative decoding with buffered verification has approximately the same runtime as the recurrent form. For $m>2$ and $d >> m$, the speedup increases linearly with $m$.

Moreover, \system reduces the per-request memory footprint during verification. 
Instead of maintaining a temporary linear attention state for each draft token, \system only buffers the draft keys and values and verifies them in parallel. 
Since state is much larger than KV, this reduction allows the serving system to support approximately $m \times$ more concurrent requests within the same memory budget. 
As a result, \system also improves overall serving throughput.

\subsection{KV-only Decoding for Short Contexts}
\label{sec:kvbuffer_parallel_decode}

As shown in Table \ref{tab:la_mem_profile}, for contexts with $d$ tokens or fewer, storing and accessing all keys and values of previous tokens are more memory-efficient than both maintaining the linear attention state with recurrent and chunkwise forms. Therefore, we prefer to compute the attention output with parallel form when context length $L < d$.

The paged memory management of \system naturally supports the parallel form, which requires to buffer keys and values of up to $d$ tokens. \system allows dynamic allocation and deallocation of KV blocks to avoid memory fragmentation, similar to paged KV cache management.

After prefilling, the serving system buffers keys and values of all tokens. While decoding, the serving system buffers the newly generated key and value and computes attention output in parallel form, as in Eq. \ref{eq:parallel_la_decoding}. Once the context length $L \geq d$, the serving system can compress the keys and values into linear attention state as Eq. \ref{eq:chunkwise_la_state_update} and then turn to chunkwise decoding as discussed above. 

Given context length of $L$, the arithmetic intensity of parallel computation form in Eq. \ref{eq:parallel_la_decoding} is $\frac{L}{L+2}$, which is close to $1$. It indicates that the parallel form is also memory-bound. Therefore, the speedup can be approximated by the ratio of average memory access between parallel and chunkwise form, given by Eq. \ref{eq:kv_only_speedup}. When $m=2\sqrt{d}$, Eq. \ref{eq:kv_only_speedup} simplifies to $\frac{d+2\sqrt{d} + 7/2}{L+2}$. The speedup decreases monotonically as $L$ increases.

\begin{align}
    \text{Speedup}_{\text{kv\_only}} (m) &\approx \frac{d+2d/m+m/2+7/2}{L+2}  \label{eq:kv_only_speedup}
\end{align}

The state size is often closely related to the memory retrieval capability of the model. However, increasing the state size can be challenging in practice because recurrent decoding must access the entire linear attention state in every decoding step, regardless of the context length. As a result, large states become prohibitively expensive, especially for short-context requests. By enabling decoding in parallel form, \system makes it feasible to scale the state size while maintaining inference efficiency for short-context decoding, further unlocking the potential of linear attention.

\subsection{Implementations}
\label{sec:implementation}

We have implemented \system in SGLang~\cite{SGLang} and Pie~\cite{Pie}. Both implementations require similar effort, with approximately 1,300 source lines of code (SLOC) in each.

\paragraph{SGLang}
We implement \system in SGLang v0.5.10 \cite{SGLang}. After the initialization of state pool, we allocate an additional memory space for \system. The buffer size and block size are configured according to the decoding scenario.
For chunkwise decoding, we first run a parameter sweep to evaluate the kernel speedup across different buffer sizes, then select the best buffer size. The block size is equal to the buffer size, so that each request fits into a single \system block. As the kernel receives a single block ID per request rather than a variable-length list of block IDs, this design simplifies the buffer management and kernel implementation.
For speculative decoding, we do not use the buffer size with best kernel speedup, instead, we set up the buffer size and block size to the number of draft tokens. In this case, the buffer of each request will be flush after draft verification. Such design make memory management simpler by avoiding dynamic block allocation during verification and supports fixed-shape parallel verification, which is compatible with CUDA Graphs.
For KV-only decoding, we use a \system block size of 16, which is a common page size in KV-cache management. As a result, each request occupies up to $\lceil L/16 \rceil$ blocks, where $L$ denotes the context length.

For chunkwise decoding and speculative decoding, since we allocate a block per-request, the total number of available blocks is the same as the number of states allocated in the state pool, which is controlled by the user-defined memory fraction reserved for linear attention state in SGLang\cite{SGLang}. For KV-only decoding, we initialize an empty state pool in SGLang because no linear-attention state will be maintained. Therefore, the total number of blocks is computed directly from the available memory, user-defined memory fraction and the block size.

\paragraph{Pie}
Pie is a programmable serving system\cite{Pie} that allows users to write inference programs, called inferlets, and submit them to the server. An inferlet manage KV-cache pages to implement different decoding algorithms. We implement \system in Pie and provide a new API, \texttt{fold}, which, when called, updates a linear attention state with selected KVs from its KV buffer.
During decoding, \system stores the inputs of recent tokens in \system, allowing state updates to be deferred. For speculative decoding, the inferlet updates the state with only accepted tokens and clears the buffered inputs of rejected tokens, avoiding storing a temporary linear attention state for every candidate token. Unlike the implementation with SGLang, where the serving system determines the timing of linear attention updates, \system in Pie is programmable: users can determine the timing of linear attention updates by calling \texttt{fold}.

\section{Experiment}

\subsection{Experimental Setup}
\label{sec:experimental_setup}

We evaluate \system on Qwen3.5 and Nemotron-3 series models. Qwen3.5 and Nemotron-3 both adopt a hybrid architecture and use Gated Delta Networks (GDN)\cite{GatedDeltaNet} and Mamba-2\cite{Mamba2} as their linear attention module, respectively. For GDN, we buffer the decay factor $\alpha$, key $\bm{k}$, and delta value $\bm{u}$ for each token to support chunkwise computation, as described in Appendix \ref{sec:gated_delta_net}. For Mamba-2, we buffer the decay factor $\alpha$, state-expansion vector $B$ and discretization-scaled input $\Delta x$ as described in Appendix \ref{sec:mamba_2}. Unless otherwise specified, the linear attention state is stored in FP32, while buffered keys and values are stored in BF16. For speculative decoding, we use Multi-Token-Prediction as the draft model for Qwen3.5-4B and evaluate on the ShareGPT dataset.

\begin{table}[t]
    \centering
    \caption{Models and serving hardware. Head dimension denotes the Mamba-2 channel width or the GDN key/value width; both Mamba-2 models have state dimension 128.}
    \label{tab:model_table}
    \begin{adjustbox}{max width=\linewidth}
    \begin{tabular}{c|c|c|c}
        \toprule
        \textbf{Model} & \shortstack{\textbf{Linear Attention} \textbf{Head Dim}} & \textbf{Precision} & \textbf{Hardware} \\
        \hline
        Nemotron-3-Nano-4B & 80 & BF16 & 1 $\times$ L40S \\
        \hline
        Nemotron-3-Super-120B-A12B & 64 & NVFP4 & 1 $\times$ B300 \\
        \hline
        Qwen3.5-4B & 128 & BF16 & 1 $\times$ L40S \\
        \hline
        Qwen3.5-122B-A10B & 128 & NVFP4 & 1 $\times$ B300 \\
        \bottomrule
    \end{tabular}
    \end{adjustbox}
\end{table}

We conduct experiments on NVIDIA L40S and B300 GPUs, with the hardware configuration for each model listed in Table~\ref{tab:model_table}. We implement \system-related kernels in Triton, including kernels for chunkwise decoding, parallel verification in speculative decoding, batched state update, and decoding in parallel form. Additional experimental results are presented in Appendix \ref{sec:additional_expr}.

\subsection{Experimental Results}
\subsubsection{Chunkwise Decoding with \system}
\label{sec:expr_chunkwise_decoding}

\begin{figure}[h]
    \centering
    \includegraphics[width=\linewidth]{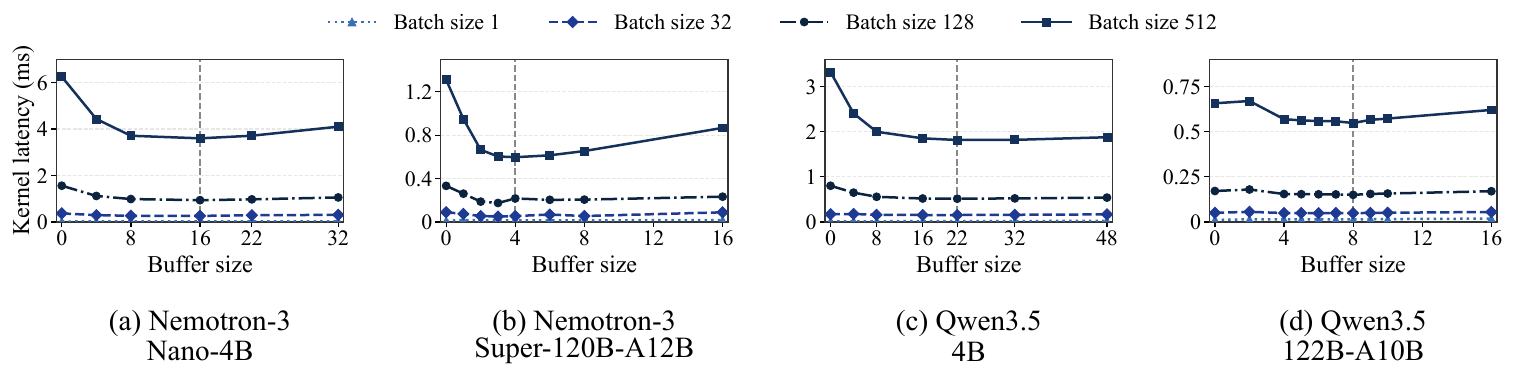}
    \caption{Kernel latency of four models with different buffer sizes $m$ and batch sizes. $m=0$ denotes recurrent decoding. For $m>0$, latency is averaged over a full buffer cycle and includes the amortized state-update cost. Vertical dashed lines mark the best tested buffer size at batch size $512$ for each model, which is used in the following kernel latency and end-to-end batch inference evaluations.}
    \label{fig:decode_kernel_buffer_sweep}
\end{figure}

\begin{figure}[htbp]
    \centering
    \begin{minipage}[t]{0.48\linewidth}
        \centering
        \includegraphics[width=\linewidth]{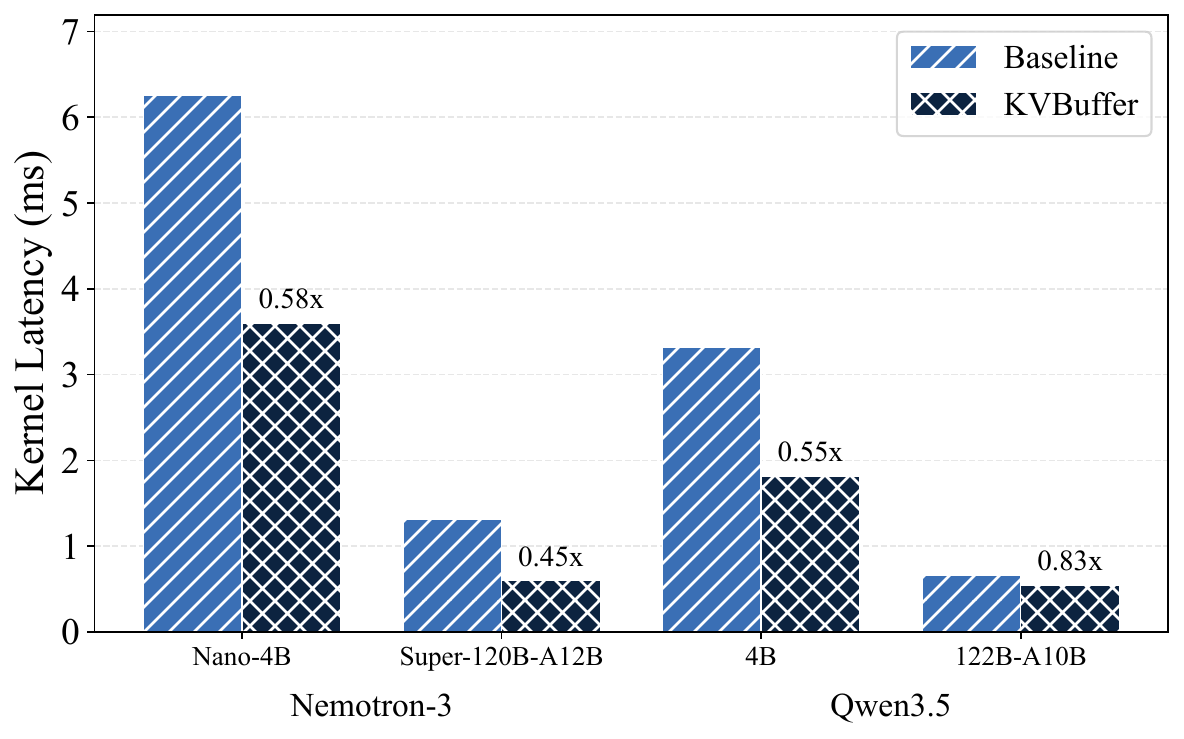}
        \caption{Kernel latency at batch size $512$, using the best tested buffer size for each model. \system latency is averaged over a full buffer cycle and includes the state-update cost.}
        \label{fig:decode_normalized_latency}
    \end{minipage}
    \hfill
    \begin{minipage}[t]{0.48\linewidth}
        \centering
        \includegraphics[width=\linewidth]{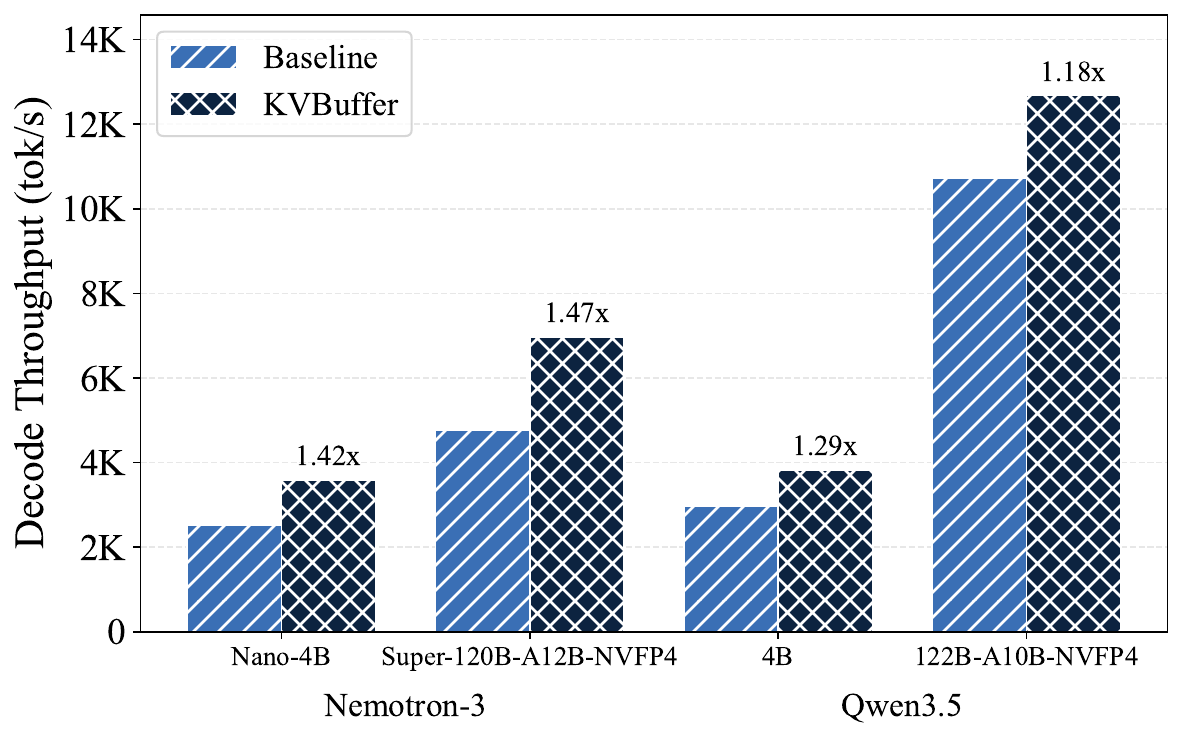}
        \caption{End-to-end batch inference decoding throughput for different models. We use a batch size of 128 for the 4B models and 512 for the 120B and 122B models.}
        \label{fig:decode_e2e_batch_throughput}
    \end{minipage}
\end{figure}

We first sweep over different buffer sizes $m$ for all four models, as shown in Figure \ref{fig:decode_kernel_buffer_sweep}. We measure kernel latency at batch sizes $1$, $32$, $128$, and $512$ for a full buffer cycle, including the amortized state-update cost. As shown in the Figure \ref{fig:decode_kernel_buffer_sweep}, latency initially decreases as state updates are amortized over more decoding steps, then increases as accessing buffered tokens becomes more expensive, consistent with the trade-off in \S\ref{sec:kvbuffer_chunkwise_decode}. The best tested buffer sizes are $16$, $4$, $22$, and $8$ for Nemotron-3-Nano, Nemotron-3-Super, Qwen3.5-4B, and Qwen3.5-122B, respectively. For Qwen3.5-4B, the best tested buffer size $m=22$ is close to $2\sqrt{128}\approx22.6$. However, Qwen3.5-122B uses the same head dimension but reaches its minimum at $m=8$ on B300. The smaller optima on B300 are consistent with the possibility discussed in \S\ref{sec:kvbuffer_chunkwise_decode} that high memory bandwidth makes flush operation compute-bound. As shown in Figure~\ref{fig:decode_normalized_latency}, chunkwise decoding with the best tested buffer sizes achieves kernel latency reductions of $0.42\times$, $0.55\times$, $0.45\times$, and $0.17\times$ the recurrent-decoding latency for the four models, respectively.

We evaluate end-to-end batch inference with an input length of $256$ tokens and an output length of $1{,}024$ tokens. For each model, we select the buffer size with the lowest kernel latency at batch size $512$. As shown in Figure~\ref{fig:decode_e2e_batch_throughput}, \system improves recurrent decoding throughput by $1.18\times$--$1.47\times$ across four models covering both GDN and Mamba-2 architectures and different model scales. Qwen3.5 achieves a smaller speedup than Nemotron-3 at similar model scales. The reason is that Qwen3.5 has a smaller linear attention state compared with Nemotron-3. Consequently, linear attention operations account for a smaller fraction of overall runtime, reducing the end-to-end benefit from the KVBuffer optimization.

\subsubsection{Parallel Verification for Speculative Decoding}

\begin{figure}[h]
    \centering
    \begin{minipage}[t]{0.48\linewidth}
        \centering
        \includegraphics[width=\linewidth]{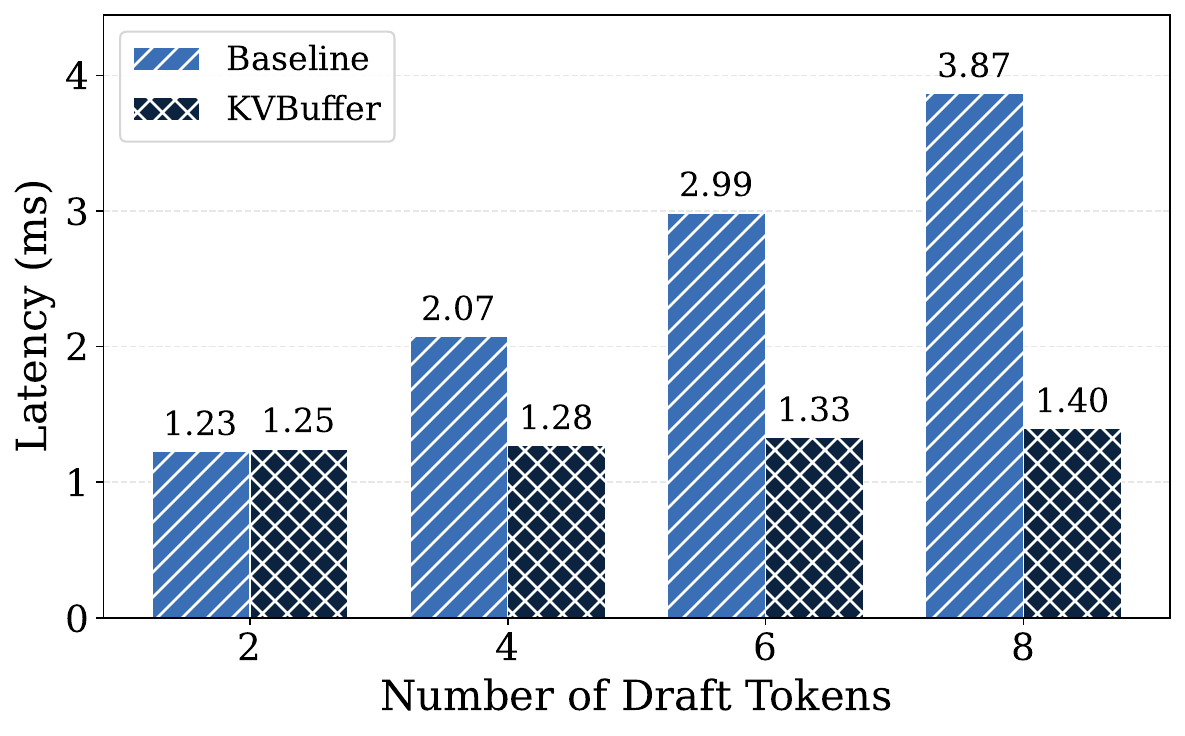}
        \caption{Kernel latency of speculative decoding verification. The latency of \system includes both attention output computation and the state update. \system incurs only a modest latency increase by verifying draft tokens in parallel.}
        \label{fig:decode_spec_latency}
    \end{minipage}
    \hfill
    \begin{minipage}[t]{0.48\linewidth}
        \centering
        \includegraphics[width=\linewidth]{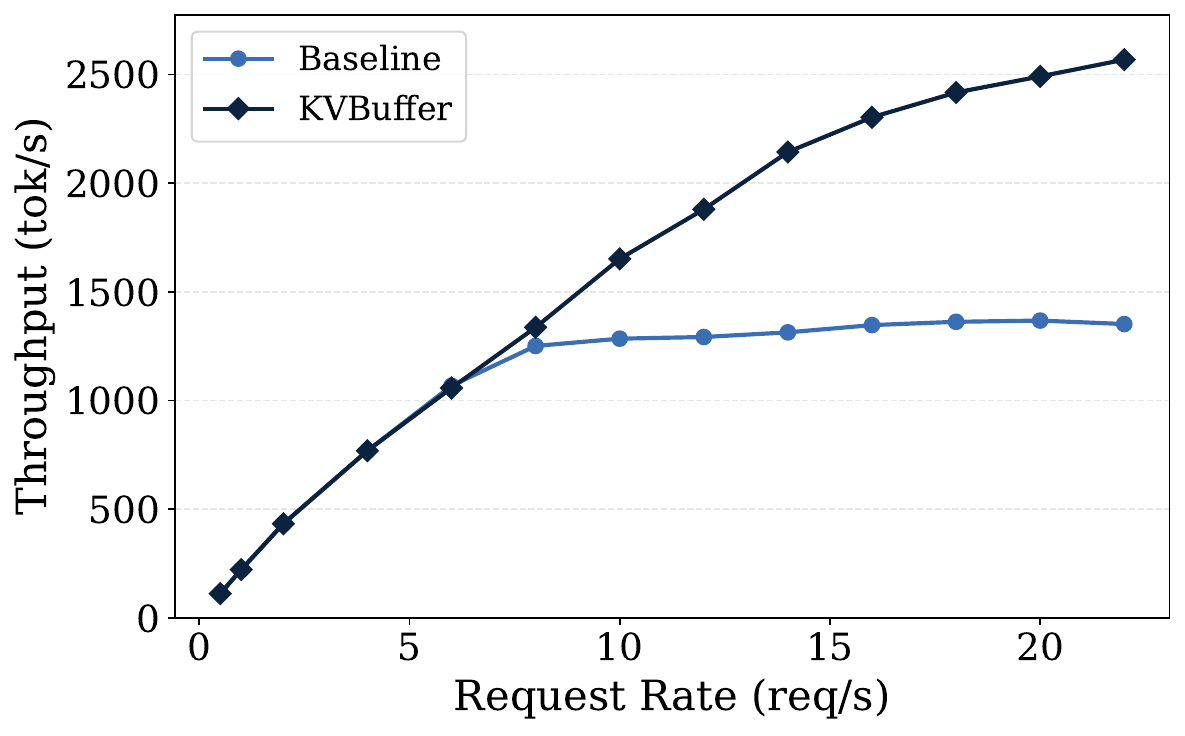}
        \caption{End-to-end serving throughput with speculative decoding. By avoiding the storage of temporary states for draft tokens, \system sustains higher request rates and improves overall throughput.}
        \label{fig:spec_throughput}
    \end{minipage}
\end{figure}

\paragraph{Parallel Verification Latency}
We next evaluate \system for speculative decoding verification on Qwen3.5-4B model. Figure~\ref{fig:decode_spec_latency} compares the kernel latency of verifying different numbers of draft tokens. For recurrent decoding, verification latency increases linearly with the number of draft tokens because the system must compute and store a temporary state for each draft token. In contrast, \system verifies draft tokens by buffering their keys and values and computing attention outputs in chunkwise form, which incurs only modest additional latency as the number of draft tokens increases. When verifying $8$ draft tokens, \system achieves a $2.77\times$ kernel speedup, closely matching the analysis in \S\ref{sec:kvbuffer_spec_decoding}, which predicts an approximately $3\times$ speedup.

\paragraph{End-to-end Throughput} 
We also evaluate the end-to-end serving throughput of \system under speculative decoding. Figure \ref{fig:spec_throughput} shows the throughput under different request rates, with the number of draft tokens set to $4$. Because recurrent verification must store a temporary state for each draft token, it supports fewer concurrent requests than chunkwise verification. As a result, \system increases the maximum number of serving requests by $5\times$ and sustains higher request rates, achieving up to a $1.9\times$ throughput improvement.

\subsubsection{KV-only Decoding for Short Contexts}

\noindent
\begin{minipage}[t]{0.46\linewidth}
    \vspace{0pt}
    Finally, we evaluate different computation forms for short-context
    requests with a batch size of $128$.
    As shown in Figure \ref{fig:decode_kv_normalized_latency}, the latency
    of KV-only decoding gradually increases with context length because
    it must read more buffered keys and values.
    As discussed in \S\ref{sec:kvbuffer_parallel_decode}, its speedup over
    chunkwise decoding decreases as the context grows; for $d=128$ and
    $m=16$, the analysis predicts a speedup of $1.0$ at $L \approx 154$.
    The measured trend is consistent with this prediction, supporting
    the advantage of KV-only decoding for short contexts, with a
    crossover near the head dimension $d$.
\end{minipage}
\hfill
\begin{minipage}[t]{0.45\linewidth}
    \vspace{0pt}
    \centering
    \includegraphics[width=\linewidth]
        {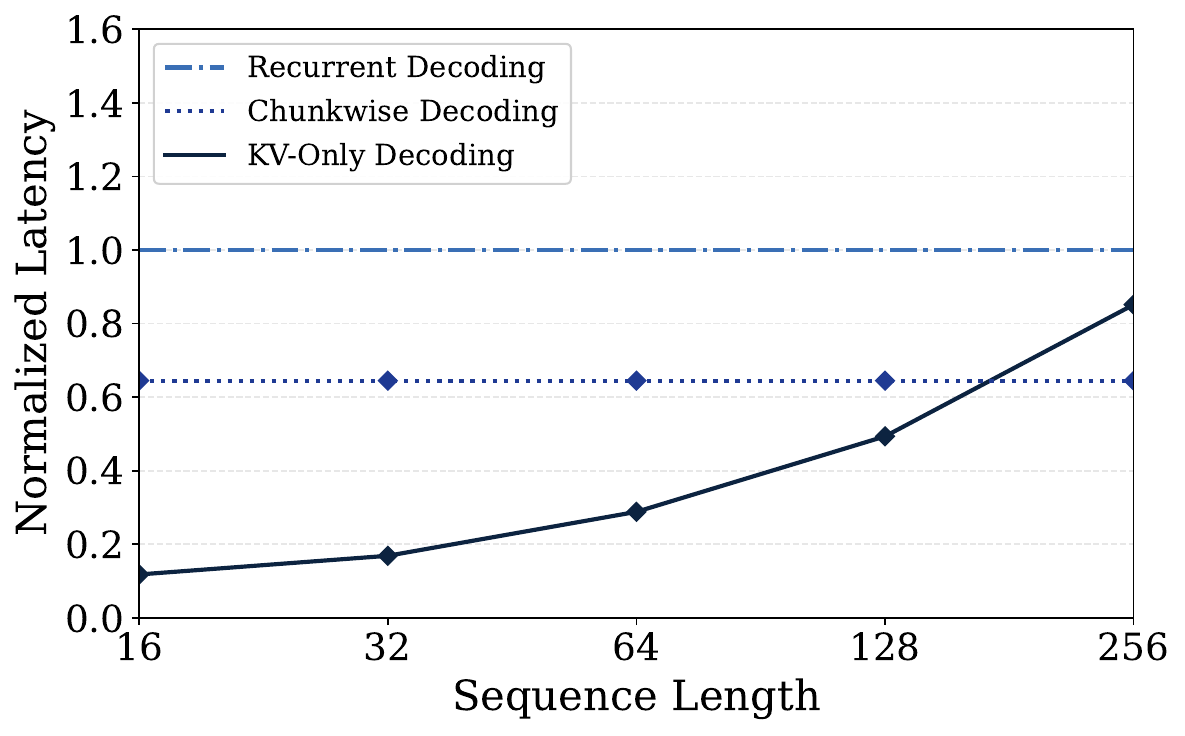}
    \captionof{figure}{Normalized kernel latency of different computation forms at batch size $128$.}
    \label{fig:decode_kv_normalized_latency}
\end{minipage}
\par\medskip

\section{Related Work}

\paragraph{Linear Attention}

Linear attention has been widely studied as an efficient alternative to softmax attention for long-context inference. 
By removing the softmax operation, linear attention reduces the quadratic dependence on sequence length and enables recurrent decoding with a fixed-size state. 
However, vanilla linear\cite{LA} attention suffers from degraded long-context retrieval and model quality compared with softmax attention. Recent variants mitigate this problem by introducing more expressive memory-control mechanisms, including data-independent decay mechanisms\cite{S4, S5, RWKV, RetNet}, data-dependent decay mechanisms\cite{Mamba, Mamba2, GLA, RWKV6} and delta rule\cite{DeltaNet, GatedDeltaNet, DeltaProduct}. These advances have made linear attention increasingly practical for real-world long-context applications.

\paragraph{Hybrid Architecture}

Recent LLMs\cite{Qwen3Next, KimiLinear, MiniMaxM1} increasingly adopt hybrid architectures that interleave linear attention layers with standard softmax-attention layers. 
These hybrid architectures preserve the strong memory retrieval capability of softmax attention while using linear attention layers to reduce the memory storage and decoding cost for long-context inference. 
Models such as Qwen3-Next\cite{Qwen3Next} and Kimi-Linear\cite{KimiLinear} demonstrate that hybrid architectures can achieve favorable quality-efficiency trade-offs, making linear attention an increasingly important building block for LLMs. 
In this work, we use Qwen3.5 and Nemotron-3 as representative hybrid architecture models for our evaluation.

\paragraph{Serving for Linear Attention}

Modern LLM serving systems, such as vLLM\cite{PagedAttention} and SGLang\cite{SGLang}, improve inference throughput through techniques such as paged KV cache management\cite{PagedAttention}, continuous batching\cite{ContinuousBatching}, prefix caching\cite{SGLang}, and speculative decoding\cite{SpecDecoding, EAGLE}. 
These techniques are primarily designed for Transformer models with softmax attention, where serving is KV cache-centric. 
In contrast, linear attention maintains a fixed-size state and no longer preserves the keys and values for all previous tokens, introducing a different memory-management problem for serving systems. 
Several recent works have begun to study serving techniques for LLMs with hybrid architectures. 
For prefix caching, Marconi\cite{Marconi} proposes efficient prefix-cache management for hybrid LLMs by checkpointing the linear attention state at appropriate positions. 
For prefill-decoding disaggregation, Prefill-as-a-Service\cite{PrefillAsService} explores cross-cluster prefill offloading for hybrid linear-attention models by exploiting their reduced memory footprint. 
For speculative decoding, STree\cite{STree} proposes a tree-based parallel verification algorithm for SSM models\cite{Mamba2, Mamba}. However, it does not investigate the linear attention models and the corresponding serving system support required for chunkwise form computation. 
In this work, we study the memory management for linear attention serving and show that it can improve memory efficiency across diverse decoding scenarios.

\section{Discussion and Limitation}
\label{sec:discussion_limitation}

\paragraph{Serving for Different Computation Forms}
\system enables linear attention to be served using different computation forms. However, dynamically selecting the most efficient form for each request is non-trivial in practice. Chunkwise decoding and KV-only decoding require different kernels and have different batching requirements, so switching computation forms based on prompt length can introduce additional scheduling overhead. 
In this work, we do not dynamically route requests across different decoding forms during serving. As hybrid models continue to scale and the hidden dimension $d$ increases, KV-only decoding may become increasingly important. One possible direction is to disaggregate short-context and long-context requests onto separate servers. Another direction is to interleave different decoding forms within the same batch, similar to chunked prefill\cite{ChunkedPrefill}. We leave the design of such dynamic scheduling mechanisms to future work.

\paragraph{Sampling} 
\system can also benefit sampling-based decoding algorithms. For example, in the beam search, the serving system needs to maintain multiple candidate branches. If each branch is served recurrently, the system must compute and store a separate linear attention state for each candidate branch, resulting in substantial memory overhead. With \system, the system can instead buffer the keys and values of candidate tokens and update the linear attention state only after the accepted branch is determined. 

\paragraph{KV-Based Prefix Caching}
Existing serving systems typically use the linear attention state as the checkpoint for prefix caching. However, because the linear attention state is so large, checkpointing is usually performed at coarse granularity, making it difficult to split or reuse prefixes at arbitrary positions. In this work, we show that the linear attention state can be reconstructed from buffered keys and values. This observation suggests an alternative prefix-caching design: instead of caching only linear attention states, serving systems may cache keys and values for selected prefixes and reconstruct the state on demand. Such a design could enable finer-grained prefix reuse and more flexible memory management for linear attention-based models.

\paragraph{Aligning Pretraining and Inference Computation}
Linear attention models are usually pretrained with chunkwise computation\cite{GLA}. However, existing serving systems typically use the recurrent form during inference, creating a mismatch between the computation form used in pretraining and that used in inference. 
For long-context generation, this mismatch can lead to accuracy degradation or instability during RL post-training\cite{DeepSeekR1}. 
Therefore, \system not only improves serving efficiency, but also helps align inference and pretraining computation, potentially improving inference stability.

\section{Conclusion}

In this paper, we present \system, an IO-aware serving mechanism for linear attention. By buffering recent keys and values, \system enables flexible computation forms across different decoding scenarios, including chunkwise decoding, speculative decoding verification, and short-context decoding. These forms reduce unnecessary memory access and substantially improve the efficiency of serving hybrid models.

\section*{Acknowledgments}
This work is supported in part by Yale University and by National Science Foundation (NSF) Athena AI Institute (Award \#2112562)

\bibliographystyle{plain}
\bibliography{abr-short,neurips_2026}


\newpage
\appendix
\section{Appendix}
\subsection{Gated Delta Networks}
\label{sec:gated_delta_net}

Gated Delta Networks (GDN)\cite{GatedDeltaNet} is among the most widely adopted variants of linear attention in recent LLMs with hybrid architecture. It incorporates a data-dependent gating mechanism $\alpha_t$, inspired by Mamba2\cite{Mamba2}, together with the delta rule\cite{DeltaNet} to selectively update the state, leading to improved performance. 

The recurrent computation form of GDN is defined as:

\begin{align}
    \tilde{\mathbf{S}}_{t-1} &= \alpha_t \mathbf{S}_{t-1} \label{eq:gated_state} \\
    \mathbf{S}_t &= \tilde{\mathbf{S}}_{t-1} - \bm{k}_t^T (\bm{k}_t \tilde{\mathbf{S}}_{t-1}) + \bm{k}_t^T (\beta_t \bm{v}_t + (1 - \beta_t) \bm{k}_t \tilde{\mathbf{S}}_{t-1}) \notag \\
    &= ((\mathbf{I} - \beta_t \bm{k}_t^T \bm{k}_t)) \tilde{\mathbf{S}}_{t-1} + \beta_t \bm{k}_t^T \bm{v}_t  \label{eq:recurrent_gdn_decoding}
\end{align}

Here, $\alpha_t$ and $\beta_t$ denote the decay factor and the learning rate at step $t$, respectively, and $\tilde{\mathbf{S}}_t$ is the gated intermediate state. Eq. \ref{eq:gated_state} implements the data-dependent gating mechanism that enables the forgetting of long-term memory. Eq. \ref{eq:recurrent_gdn_decoding} corresponds to the delta rule. Specifically, it removes the retrieved old value $\bm{k}_t \tilde{\mathbf{S}}_{t-1}$ from gated state $\tilde{\mathbf{S}}_{t-1}$ and replaces it with a new combination value $(\beta_t \bm{v}_t + (1 - \beta_t) \bm{k}_t \tilde{\mathbf{S}}_{t-1})$. This formulation makes explicit that the update consists of erasing stale information and writing new content in a controlled manner.

As in linear attention, the recurrent form of GDN maintains a fixed-size state $\mathbf{S}_t$, resulting in constant computational and memory cost with respect to context length. Compared to linear attention, GDN introduces additional gating and updating operations, which enhance expressivity and allow more flexible control over memory updates, leading to substantially better model quality than linear attention.

GDN can also be written in a parallel form:

\begin{align}
    \gamma_t &= \prod_{i \in \text{Par}(t)}{\alpha_i} \notag \\
    \mathbf{\Gamma}_{ij} &= \frac{\gamma_i}{\gamma_j}, i \geq j \text{ and } \mathbf{M}_{ij} \neq 0, \text{otherwise 0} \notag \\
    \mathbf{A} &= [\mathbf{I} + \text{strictLower}(\text{Diag}(\bm{\beta})(\mathbf{\Gamma} \odot \mathbf{K}^T \mathbf{K}))]^{-1} \notag \\
    \tilde{\mathbf{K}} &= \text{Diag}(\bm{\gamma}) \mathbf{A} \text{Diag}(\bm{\beta}) \mathbf{K}; \tilde{\mathbf{V}} = \mathbf{A} \text{Diag}(\bm{\beta}) \mathbf{V} \notag \\
    \mathbf{O} &= (\mathbf{Q}\mathbf{K}^T \odot \mathbf{\Gamma}) \tilde{\mathbf{V}} \label{eq:parallel_gdn_decoding}
\end{align}

At the single-token level, this parallel formulation can be expressed as:

\begin{align}
    \bm{u}_t &= \beta_t \bm{v}_t - \beta_t (\Sigma_{i=1}^{j}{\frac{\gamma_t} {\gamma_{t-i}} \bm{k}_t \bm{k}_{t-i}^T \bm{u}_i}) \notag \\
    \bm{o}_t &= \Sigma_{i=0}^{j}{\frac{\gamma_t}{\gamma_{t-i}} \bm{q}_t \bm{k}_{t-i}^T \bm{u}_i} \label{eq:parallel_gdn_decoding}
\end{align}

where $\text{Par}(t)$ denotes the set of prefix tokens, including the token $t$, and $\mathbf{\Gamma}$ is the gated mask encoding the cumulative effect of $\alpha$.

In addition to the recurrent and parallel forms, GDN also supports a chunkwise computation form, as shown in Eq. \ref{eq:chunkwise_gdn_spec_verify}:

\begin{align}
    \gamma_t &= \prod_{i \in \text{Par}(t)}{\alpha_i} \notag \\
    \mathbf{\Gamma}_{ij} &= \frac{\gamma_i}{\gamma_j}, i \geq j \text{ and } \mathbf{M}_{ij} \neq 0, \text{otherwise 0} \notag \\
    \mathbf{A} &= [\mathbf{I} + \text{strictLower}(\text{Diag}(\bm{\beta})(\mathbf{\Gamma} \odot \mathbf{K}^T \mathbf{K}))]^{-1} \notag \\
    \tilde{\mathbf{K}} &= \text{Diag}(\bm{\gamma}) \mathbf{A} \text{Diag}(\bm{\beta}) \mathbf{K}; \tilde{\mathbf{V}} = \mathbf{A} \text{Diag}(\bm{\beta}) \mathbf{V} \notag \\
    \mathbf{U} &= \tilde{\mathbf{V}} - \tilde{\mathbf{K}} \mathbf{S}_{t-j-1} \notag \\
    \mathbf{O} &= \text{Diag}(\bm{\gamma}) \mathbf{Q} \mathbf{S}_{t-j-1} + (\mathbf{Q}\mathbf{K}^T \odot \mathbf{\Gamma}) \mathbf{U} \label{eq:chunkwise_gdn_spec_verify}
\end{align}

The corresponding single-token chunkwise computation is:

\begin{align}
    \gamma_t &= \prod_{i=t-m}^{t} \alpha_i \notag \\
    \bm{u}_t &= \beta_t \bm{v}_t - \beta_t (\gamma_t \bm{k}_t \mathbf{S}_{t-j-1} + \Sigma_{i=1}^{j}{\frac{\gamma_t} {\gamma_{t-i}} \bm{k}_t \bm{k}_{t-i}^T \bm{u}_i}) \notag \\
    \bm{o}_t &= \gamma_t \bm{q}_t \mathbf{S}_{t-j-1} + \Sigma_{i=0}^{j}{\frac{\gamma_t}{\gamma_{t-i}} \bm{q}_t \bm{k}_{t-i}^T \bm{u}_i} \notag \\
    \mathbf{S}_t &= \gamma_t \mathbf{S}_{t-j-1} + \Sigma_{i=0}^{j}{\frac{\gamma_t}{\gamma_{t-i}}  \bm{k}_i^T \bm{u}_i} \label{eq:chunkwise_gdn_decoding}
\end{align}

For GDN, we need to buffer the decay factor $\alpha_t$, key $\bm{k}_t$ and delta value $\bm{u}_t$. Since $\alpha_t$ is a scalar, the additional memory overhead is small. As shown in Table \ref{tab:gdn_mem_profile}, the overall storage and memory access of GDN remain close to that of linear attention.

\begin{table}[htbp]
    \centering
    \caption{Memory storage and average per-token memory access during decoding for different forms of GDN. $L$ denotes the context length, $d$ the hidden dimension, and $m$ the chunk size. Memory access is measured in bytes, assuming that the linear attention state is stored in FP32 and $\bm{q}, \bm{k}, \bm{v}, \bm{o}$ are stored in FP16. For the recurrent form, the state update and attention output computation are fused into a single kernel.}
    \resizebox{\linewidth}{!}{
        \begin{tabular}{c|c|c|c}
            \toprule
            \multirow{2}{*}{Computation Forms} & \multirow{2}{*}{Memory Storage} & \multicolumn{2}{c}{Memory Access} \\
            \cline{3-4}
            & & Read & Write \\
            \hline
            Parallel & $4Ld + 2L$ & $4Ld + 2d + 2L + 2$ & $6d + 2$ \\
            \hline
            Recurrent & $4d^2$ & $4d^2 + 6d + 4$ & $4d^2 + 2d$ \\
            \hline
            Chunkwise & $4d^2 + 4md + 2m$ & $4(1+1/m) d^2 + 2(m+4)d + m + 5$ & $4d^2 / m + 6d + 2$ \\
            \bottomrule
        \end{tabular}
    }
    \label{tab:gdn_mem_profile}
\end{table}

Furthermore, based on the memory-access estimates in Table \ref{tab:gdn_mem_profile}, the estimated speedups of GDN for chunkwise decoding, speculative-decoding verification, and KV-only decoding are as follows:

\begin{align}
    \text{Speedup}_\text{gdn\_chunkwise\_decoding} &= \frac{8d^2+8d+4}{4d^2+\frac{8d^2}{m}+2md+14d+m+7} \notag \\
    &\approx \frac{4(d+1)}{2d+\frac{4d}{m}+m+7}
\end{align}

\begin{align}
    \text{Speedup}_\text{gdn\_parallel\_verify} &= \frac{4(m+1)d^2+8md+4m}{12d^2+16md+8m} \notag \\
    &\approx \frac{(m+1)d+2m}{3d+4m}
\end{align}

\begin{align}
    \text{Speedup}_\text{gdn\_kv\_only} &= \frac{4d^2+\frac{8d^2}{m}+2md+14d+m+7}{4Ld+8d+2L+4} \notag \\
    &\approx \frac{d+2d/m+m/2+7/2}{L+2}
\end{align}

These estimates show that the speedups of GDN are close to those of linear attention. This is because GDN only requires buffering additional scalar decay factors, while the dominant storage and memory-access costs remain determined by the key, value, and state tensors.

\subsection{Mamba-2}
\label{sec:mamba_2}

For simplicity, we consider one Mamba-2 head and omit the head index. In Mamba-2, the state-transition matrix has the scalar-times-identity form $A_t=a_tI$. Therefore, we use the scalar $a_t$ in place of $A_t$ for simplicity.

Let $x_t$ denote the token representation, $B_t$ the state-expansion vector, $C_t$ the state-contraction vector, $a_t$ the discrete state-decay factor, and $S_t$ the Mamba-2 state. Mamba-2 recurrently updates its state and computes its output $o_t$ as:

\begin{align}
    S_t = a_t S_{t-1} + B_t x_t^T
\end{align}

\begin{align}
    o_t = C_t^T S_t
\end{align}

Define the cumulative decay from position $s$ to position $u$ as $\lambda_{t \leftarrow s} = \prod_{r=s+1}^{t}{a_r}$. Note that $\lambda_{s \leftarrow s}=1$.

By unrolling the recurrence,  we can derive the chunkwise computation form as:

\begin{align}
    o_j = \lambda_{j \leftarrow t-m} C_j^T S_{t-m} + \sum_{i=t-m+1}^{j}{\lambda_{j \leftarrow i} (C_j^T B_i) x_i}, \text{for} j=t-m+1,...,t
\end{align}

\begin{align}
    S_t = \lambda_{t \leftarrow t-m} S_{t-m} + \sum_{i=t-m+1}^{t}{\lambda_{t \leftarrow i} B_i x_i^T}
\end{align}

Define the lower-triangular decay matrix:

\begin{align}
    \Gamma_{ij} = \lambda_{i \leftarrow j}, i \geq j, \text{otherwise } 0
\end{align}

and stack the token vectors row-wise into $X, B, C$, we can derive the parallel form as:

\begin{align}
    O = (CB^T \odot \Gamma) X
\end{align}

Therefore, Mamba-2 can support \system for the three scenarios we discussed in the paper.

\clearpage
\subsection{Additional Experiment}
\label{sec:additional_expr}

\subsubsection{End-to-end Serving Performance}

We evaluate end-to-end online serving performance for standard decoding using requests sampled from the ShareGPT dataset. As shown in Table \ref{tab:e2e_serving_perf}, \system provides increasingly clear benefits as the request rate and decode concurrency increase. At moderate to high request rates, it improves throughput by up to 10\%, reduces mean end-to-end latency by up to 18\%, and reduces mean time per output token by up to 20\%.

\noindent\begin{minipage}{\linewidth}
\centering
\captionof{table}{End-to-end standard decoding serving performance on ShareGPT.
Each entry is reported as ``\system / baseline'', followed by their ratio in parentheses.}
\label{tab:e2e_serving_perf}

\resizebox{\textwidth}{!}{
\begin{tabular}{lccccc}
\toprule
\textbf{Model}
& \textbf{Request Rate (req/s)}
& \textbf{Throughput (tok/s)} 
& \textbf{E2E (ms)}
& \textbf{TTFT (ms)}
& \textbf{TPOT (ms)} \\
\midrule

\multirow{3}{*}{Nemotron-3-Nano-4B}
& 1
& 219.4/219.4 (\textbf{1.00$\times$})
& 2924.1/2979.2 (\textbf{0.98$\times$})
& 48.6/49.4 (\textbf{0.98$\times$})
& 12.91/13.16 (\textbf{0.98$\times$}) \\
& 8
& 1066.2/997.0 (\textbf{1.07$\times$})
& 5117.9/6104.5 (\textbf{0.84$\times$})
& 63.7/69.3 (\textbf{0.92$\times$})
& 23.94/28.89 (\textbf{0.83$\times$}) \\
& 16
& 1206.6/1099.4 (\textbf{1.10$\times$})
& 6980.9/8515.2 (\textbf{0.82$\times$})
& 98.0/112.6 (\textbf{0.87$\times$})
& 36.96/46.32 (\textbf{0.80$\times$}) \\

\midrule

\multirow{3}{*}{Nemotron-3-Super-120B-A12B}
& 1
& 217.8/218.2 (\textbf{1.00$\times$})
& 4183.8/4081.1 (\textbf{1.03$\times$})
& 114.9/112.3 (\textbf{1.02$\times$})
& 19.95/19.59 (\textbf{1.02$\times$}) \\
& 8
& 818.0/788.7 (\textbf{1.04$\times$})
& 11335.7/12247.5 (\textbf{0.93$\times$})
& 211.4/222.7 (\textbf{0.95$\times$})
& 69.42/75.94 (\textbf{0.91$\times$}) \\
& 16
& 941.9/901.4 (\textbf{1.04$\times$})
& 12122.7/13186.9 (\textbf{0.92$\times$})
& 251.3/260.6 (\textbf{0.96$\times$})
& 130.42/145.50 (\textbf{0.90$\times$}) \\

\midrule

\multirow{3}{*}{Qwen3.5-4B}
& 1
& 219.4/219.5 (\textbf{1.00$\times$})
& 3553.4/3529.5 (\textbf{1.01$\times$})
& 60.5/63.3 (\textbf{0.96$\times$})
& 16.20/16.21 (\textbf{1.00$\times$}) \\
& 8
& 1019.5/985.5 (\textbf{1.03$\times$})
& 5809.5/6345.3 (\textbf{0.92$\times$})
& 53.8/55.3 (\textbf{0.97$\times$})
& 28.56/31.29 (\textbf{0.91$\times$}) \\
& 16
& 1141.0/1080.9 (\textbf{1.06$\times$})
& 7249.8/8256.8 (\textbf{0.88$\times$})
& 64.7/68.8 (\textbf{0.94$\times$})
& 41.76/48.05 (\textbf{0.87$\times$}) \\

\midrule

\multirow{3}{*}{Qwen3.5-122B-A10B}
& 1
& 223.2/223.3 (\textbf{1.00$\times$})
& 1744.5/1679.7 (\textbf{1.04$\times$})
& 99.7/99.7 (\textbf{1.00$\times$})
& 8.48/8.16 (\textbf{1.04$\times$}) \\
& 8
& 1223.0/1228.6 (\textbf{1.00$\times$})
& 6144.6/6218.2 (\textbf{0.99$\times$})
& 129.1/123.6 (\textbf{1.04$\times$})
& 35.62/36.17 (\textbf{0.98$\times$}) \\
& 16
& 1503.4/1521.0 (\textbf{0.99$\times$})
& 7844.4/7919.9 (\textbf{0.99$\times$})
& 134.2/130.4 (\textbf{1.03$\times$})
& 66.62/72.14 (\textbf{0.92$\times$}) \\

\bottomrule
\end{tabular}
}
\end{minipage}

\subsubsection{Numerical Stability and Accuracy}

We evaluate kernel-level numerical equivalence using Gaussian random inputs for Qwen3-Next deployed across four NVIDIA L40S GPUs. We run up to 1,024 decoding steps with the same initial state and token inputs for both methods. Table \ref{tab:kernel_numerical_diff} demonstrates the mean element-wise absolute error, mean relative error, and cosine similarity between the outputs of chunkwise and recurrent decoding. The results show that chunkwise decoding remains numerically stable and produces outputs that are very close to those of recurrent decoding over long decoding steps. 
We also evaluate the numerical equivalence for the linear attention state tensor. Table \ref{tab:state_numerical_diff} show that the state tensors also remain highly similar across long decoding steps.

\noindent\begin{minipage}{\linewidth}
\centering
\captionof{table}{Numerical differences between output vectors produced by
chunkwise and recurrent decoding at different decode lengths for Qwen3-Next.
The buffer size is set to 16.}
\label{tab:kernel_numerical_diff}

\resizebox{\textwidth}{!}{
\begin{tabular}{lcccccccccc}
\toprule
\textbf{Decode Length}
& \textbf{1}
& \textbf{4}
& \textbf{8}
& \textbf{16}
& \textbf{32}
& \textbf{64}
& \textbf{128}
& \textbf{256}
& \textbf{512}
& \textbf{1024} \\
\midrule

Abs. Error ($\times 10^{-8}$)
& 0.1295
& 4.892
& 3.997
& 4.000
& 4.614
& 4.220
& 3.981
& 3.511
& 4.924
& 3.682 \\

Rel. Error ($\times 10^{-3}$)
& 0.00057
& 0.4801
& 0.5462
& 0.8494
& 1.300
& 0.8206
& 0.8707
& 0.7215
& 2.371
& 0.5063 \\

Cos. Distance ($\times 10^{-8}$)
& 0.010
& 2.120
& 2.820
& 2.350
& 4.100
& 2.470
& 2.190
& 2.470
& 3.700
& 2.410 \\

\bottomrule
\end{tabular}
}

\end{minipage}

\par\addvspace{6pt}
\noindent\begin{minipage}{\linewidth}
\centering
\captionof{table}{Numerical differences between the linear attention states produced by
chunkwise and recurrent decoding for Qwen3-Next. The buffer size is set to 16.}
\label{tab:state_numerical_diff}

\begin{tabular}{lccccccc}
\toprule
\textbf{Metric}
& \textbf{16}
& \textbf{32}
& \textbf{64}
& \textbf{128}
& \textbf{256}
& \textbf{512}
& \textbf{1024} \\
\midrule

Abs. Error ($\times 10^{-7}$)
& 9.446
& 9.171
& 9.396
& 9.120
& 9.252
& 9.350
& 9.162 \\

Rel. Error ($\times 10^{-3}$)
& 1.304
& 1.371
& 1.520
& 1.239
& 1.212
& 1.295
& 1.830 \\

Cos. Distance ($\times 10^{-8}$)
& 4.360
& 4.320
& 4.130
& 4.240
& 4.380
& 4.250
& 4.320 \\

\bottomrule
\end{tabular}

\end{minipage}
\par\addvspace{6pt}

Parallel decoding exhibits similar numerical stability. As shown in Table \ref{tab:parallel_output_numerical_diff} and Table~\ref{tab:parallel_state_numerical_diff}, its linear attention outputs and states closely match those of recurrent decoding, with errors remaining stable across all tested decoding lengths.

\noindent\begin{minipage}{\linewidth}
\centering
\captionof{table}{Numerical differences between output vectors produced by
parallel and recurrent decoding at different decode lengths for Qwen3-Next.
The buffer size is set to 16.}
\label{tab:parallel_output_numerical_diff}

\resizebox{\textwidth}{!}{
\begin{tabular}{lcccccccccc}
\toprule
\textbf{Metric}
& \textbf{1}
& \textbf{4}
& \textbf{8}
& \textbf{16}
& \textbf{32}
& \textbf{64}
& \textbf{128}
& \textbf{256}
& \textbf{512}
& \textbf{1024} \\
\midrule

Abs. Error ($\times 10^{-8}$)
& 0.03104
& 4.477
& 4.446
& 4.428
& 4.486
& 4.570
& 4.454
& 4.414
& 4.496
& 4.417 \\

Rel. Error ($\times 10^{-3}$)
& 0.0003971
& 0.6138
& 1.114
& 1.061
& 0.9969
& 1.277
& 1.170
& 1.224
& 1.135
& 1.046 \\

Cos. Distance ($\times 10^{-8}$)
& 0.000
& 1.970
& 3.170
& 2.970
& 3.090
& 3.190
& 3.020
& 3.040
& 3.010
& 2.930 \\

\bottomrule
\end{tabular}
}
\end{minipage}

\par\addvspace{6pt}
\noindent\begin{minipage}{\linewidth}
\centering
\captionof{table}{Numerical differences between the linear attention states produced by
parallel and recurrent decoding  for Qwen3-Next.
The buffer size is set to 16.}
\label{tab:parallel_state_numerical_diff}

\begin{tabular}{lccccccc}
\toprule
\textbf{Metric}
& \textbf{16}
& \textbf{32}
& \textbf{64}
& \textbf{128}
& \textbf{256}
& \textbf{512}
& \textbf{1024} \\
\midrule

Abs. Error ($\times 10^{-7}$)
& 9.194
& 9.178
& 9.238
& 9.222
& 9.215
& 9.209
& 9.223 \\

Rel. Error ($\times 10^{-3}$)
& 1.370
& 1.389
& 1.388
& 1.338
& 1.376
& 1.336
& 1.372 \\

Cos. Distance ($\times 10^{-8}$)
& 4.280
& 4.280
& 4.300
& 4.300
& 4.300
& 4.290
& 4.280 \\

\bottomrule
\end{tabular}
\end{minipage}
\par\addvspace{6pt}

Furthermore we evaluate the downstream task accuracy on 1,000 GSM8K questions using the 5-shot setting for chunkwise decoding. GSM8K usually has tens of generated tokens per question and uses exact-match accuracy measurement. The results in Table \ref{tab:gsm8k_acc} show that chunkwise decoding with various buffer sizes has close accuracy with recurrent decoding.

\noindent\begin{minipage}{\linewidth}
    \centering
    \captionof{table}{GSM8K accuracy for Qwen3-Next serving with different buffer sizes. A buffer size of 0 denotes the recurrent decoding.}
    \label{tab:gsm8k_acc}
    \begin{tabular}{l|c|c|c|c|c|c}
        \toprule
        \textbf{Buffer Size} & \textbf{0} & \textbf{4} & \textbf{8} & \textbf{16} & \textbf{32} & \textbf{64} \\
        \hline
        Chunkwise Decoding & 0.945 & 0.949 & 0.943 & 0.941 & 0.941 & 0.945 \\
        Parallel Decoding & 0.945 & 0.945 & 0.941 & 0.948 & 0.946 & 0.948 \\
        \bottomrule
    \end{tabular}
\end{minipage}

Finally, we examine whether small numerical differences affect speculative decoding by comparing the average number of accepted tokens under parallel and recurrent verification. On Qwen3-Next, both the baseline and \system achieve an average of 2.99 accepted tokens, indicating that \system does not noticeably affect token acceptance.

\subsubsection{Breakdown}

We profile end-to-end batched decoding for Nemotron-3-Nano-4B on an L40S GPU with buffer size of 16. Table \ref{tab:kernel_time_breakdown} reports the GPU kernel time for each operation category, averaged over 64 decoding steps. The recurrent SSM update accounts for 63.9\% of the baseline GPU runtime. \system replaces this operation with buffered decoding and periodic state flushing, reducing the average SSM-related runtime from 31.409 ms to 20.429 ms, corresponding to a 1.537x speedup for the core SSM computation. This reduction translates into a 1.314x end-to-end decoding speedup.

\par\addvspace{6pt}
\noindent\begin{minipage}{\linewidth}
\centering
\captionof{table}{Breakdown of GPU kernel time per decode step for end-to-end batched
inference with Nemotron-3-Nano-4B on an L40S GPU. Times are averaged over
64 decoding steps. ``SSM update'' includes only the core SSM operations,
while the associated projections are categorized as GEMMs.}
\label{tab:kernel_time_breakdown}

\begin{tabular}{lccccc}
\toprule
\multirow{2}{*}{\textbf{Category}}
& \multicolumn{2}{c}{\textbf{Baseline}}
& \multicolumn{2}{c}{\textbf{\system}}
& \multirow{2}{*}{\textbf{Speedup}} \\
\cmidrule(lr){2-3}
\cmidrule(lr){4-5}
& \textbf{Time (ms)}
& \textbf{Ratio}
& \textbf{Time (ms)}
& \textbf{Ratio}
& \\
\midrule

SSM Update
& 31.409
& 63.9\%
& 0
& 0.0\%
& -- \\

\system Decode
& 0
& 0.0\%
& 17.986
& 48.1\%
& -- \\

\system Flush
& 0
& 0.0\%
& 2.443
& 6.5\%
& -- \\

Conv
& 0.328
& 0.7\%
& 0.310
& 0.8\%
& -- \\

GEMM
& 13.477
& 27.4\%
& 13.130
& 35.1\%
& -- \\

Normalization
& 1.099
& 2.2\%
& 0.805
& 2.2\%
& -- \\

Other
& 2.820
& 5.7\%
& 2.710
& 7.2\%
& -- \\

\midrule

Overall SSM Operations
& 31.409
& 63.9\%
& 20.429
& 54.6\%
& \textbf{1.537$\times$} \\

\textbf{Overall}
& \textbf{49.132}
& \textbf{100\%}
& \textbf{37.384}
& \textbf{100\%}
& \textbf{1.314$\times$} \\

\bottomrule
\end{tabular}
\end{minipage}
\par\addvspace{6pt}

\clearpage
\subsubsection{Per-step Decoding Latency}

As shown in Figure \ref{fig:perstep_latency}, we measure the end-to-end latency of 16 consecutive decoding steps for Nemotron-3-Nano-4B on an L40S GPU with a buffer size of 16. As the buffer fills, \system decoding latency gradually increases because each step reads and processes more buffered keys and values. When the buffer becomes full at step 15, the buffer is flushed, producing a periodic latency spike.

\par\addvspace{6pt}
\noindent\begin{minipage}{\linewidth}
    \centering
    \includegraphics[width=0.5\linewidth]{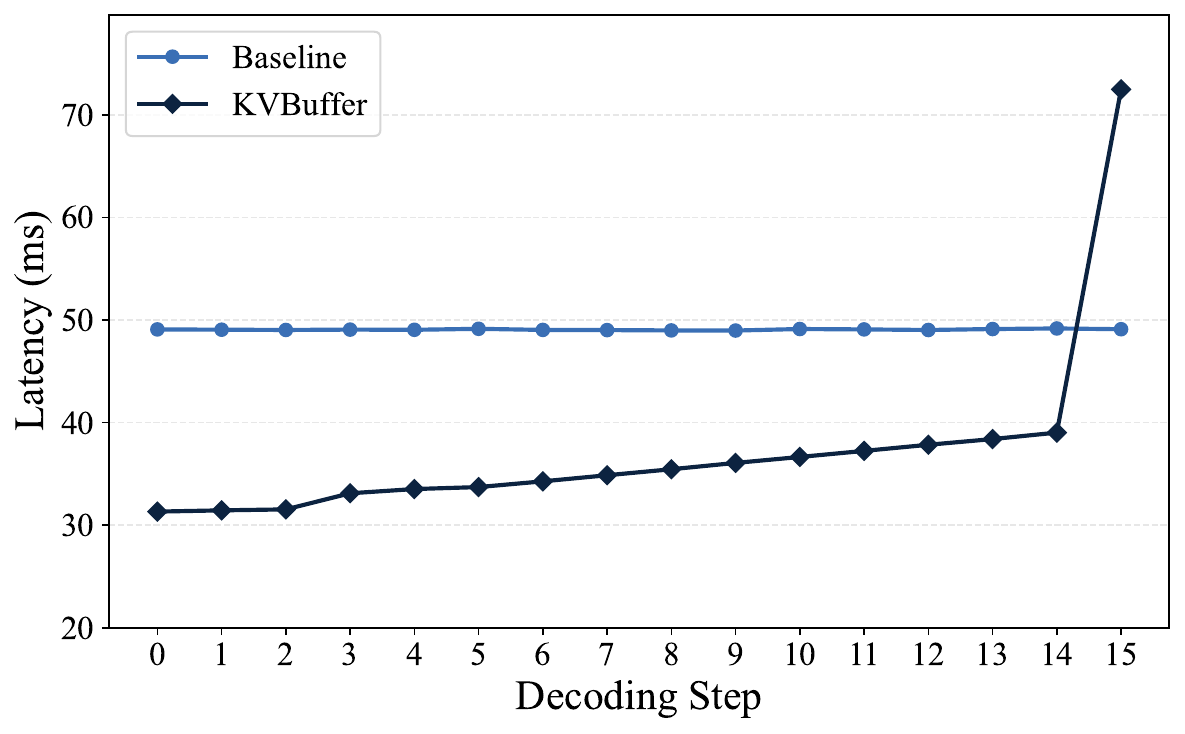}
    \captionof{figure}{Per-step Decoding Latency}
    \label{fig:perstep_latency}
\end{minipage}
\par\addvspace{6pt}

\subsubsection{Memory Overhead}

Table \ref{tab:kvbuffer_memory_overhead} reports the additional memory required to store the buffered keys and values for Qwen3-Next model. With a buffer size of 22, which we use for standard chunkwise decoding, \system requires 2.344 MB per request per GPU. This corresponds to approximately 12.7\% of the original per-request linear-attention state memory. Because the maximum number of concurrent requests is determined by the capacity of the state-memory pool, adding a buffer of size 22 reduces the theoretical maximum request capacity by approximately 11.3\%.

For speculative decoding with four draft tokens, \system introduces only 0.426 MB of additional memory per request per GPU, corresponding to a 2.3\% memory overhead and an approximately 2.26\% reduction in theoretical request capacity. Therefore, its memory impact is small in the typical speculative-decoding setting.

\par\addvspace{6pt}
\noindent\begin{minipage}{\linewidth}
\centering
\captionof{table}{Per-request memory overhead of \system at different buffer sizes.
The measurements are obtained when serving Qwen3-Next on four L40S GPUs.
All values are reported per request and per GPU.}
\label{tab:kvbuffer_memory_overhead}

\begin{tabular}{ccc}
\toprule
\textbf{Buffer Size}
& \textbf{KVBuffer}
& \textbf{Linear Attention State} \\
& \textbf{(MB)}
& \textbf{(MB)} \\
\midrule

4  & 0.426 & 18.42 \\
8  & 0.853 & 18.42 \\
16 & 1.705 & 18.42 \\
22 & 2.344 & 18.42 \\
32 & 3.410 & 18.42 \\

\bottomrule
\end{tabular}
\end{minipage}
\par\addvspace{6pt}


\newpage
\section*{NeurIPS Paper Checklist}

\begin{enumerate}

\item {\bf Claims}
    \item[] Question: Do the main claims made in the abstract and introduction accurately reflect the paper's contributions and scope?
    \item[] Answer: \answerYes{} 
    \item[] Justification: The main claims in the abstract and introduction reflect the contributions and scope accurately.

\item {\bf Limitations}
    \item[] Question: Does the paper discuss the limitations of the work performed by the authors?
    \item[] Answer: \answerYes{} 
    \item[] Justification: We discuss the limitations of this work in \S\ref{sec:discussion_limitation}.

\item {\bf Theory assumptions and proofs}
    \item[] Question: For each theoretical result, does the paper provide the full set of assumptions and a complete (and correct) proof?
    \item[] Answer: \answerYes{} 
    \item[] Justification: We state the assumptions for each theoretical result and provide proof sketches.

    \item {\bf Experimental result reproducibility}
    \item[] Question: Does the paper fully disclose all the information needed to reproduce the main experimental results of the paper to the extent that it affects the main claims and/or conclusions of the paper (regardless of whether the code and data are provided or not)?
    \item[] Answer: \answerYes{} 
    \item[] Justification: We disclose all details needed to reproduce the experimental results.

\item {\bf Open access to data and code}
    \item[] Question: Does the paper provide open access to the data and code, with sufficient instructions to faithfully reproduce the main experimental results, as described in supplemental material?
    \item[] Answer: \answerNo{} 
    \item[] Justification: We have not released the code at submission time because the algorithm is simple to implement, and our current implementation is built on SGLang and still requires additional engineering before release.

\item {\bf Experimental setting/details}
    \item[] Question: Does the paper specify all the training and test details (e.g., data splits, hyperparameters, how they were chosen, type of optimizer) necessary to understand the results?
    \item[] Answer: \answerYes{} 
    \item[] Justification: We provide experimental setting in \S\ref{sec:experimental_setup}.

\item {\bf Experiment statistical significance}
    \item[] Question: Does the paper report error bars suitably and correctly defined or other appropriate information about the statistical significance of the experiments?
    \item[] Answer: \answerNo{} 
    \item[] Justification: For each kernel-level experiment, we perform a warm-up phase and then report the average over 10 repeated runs; since the measured results are stable across runs, we omit error bars for clarity.

\item {\bf Experiments compute resources}
    \item[] Question: For each experiment, does the paper provide sufficient information on the computer resources (type of compute workers, memory, time of execution) needed to reproduce the experiments?
    \item[] Answer: \answerYes{} 
    \item[] Justification: We provide details of computer resources in \S\ref{sec:experimental_setup}.
    
\item {\bf Code of ethics}
    \item[] Question: Does the research conducted in the paper conform, in every respect, with the NeurIPS Code of Ethics \url{https://neurips.cc/public/EthicsGuidelines}?
    \item[] Answer: \answerYes{} 
    \item[] Justification: This research is conducted in accordance with the NeurIPS Code of Ethics.

\item {\bf Broader impacts}
    \item[] Question: Does the paper discuss both potential positive societal impacts and negative societal impacts of the work performed?
    \item[] Answer: \answerNA{} 
    \item[] Justification: This work focuses on system-level efficiency improvements for serving linear-attention LLMs and does not introduce new models, datasets, or deployment-specific applications, so it has no direct societal impact beyond general inference-efficiency gains.
    
\item {\bf Safeguards}
    \item[] Question: Does the paper describe safeguards that have been put in place for responsible release of data or models that have a high risk for misuse (e.g., pre-trained language models, image generators, or scraped datasets)?
    \item[] Answer: \answerNA{} 
    \item[] Justification: This paper does not release new models, datasets, or other assets with high risk for misuse, so no additional safeguards are required.

\item {\bf Licenses for existing assets}
    \item[] Question: Are the creators or original owners of assets (e.g., code, data, models), used in the paper, properly credited and are the license and terms of use explicitly mentioned and properly respected?
    \item[] Answer: \answerYes{} 
    \item[] Justification: We credit and mention the assets usage in the paper.

\item {\bf New assets}
    \item[] Question: Are new assets introduced in the paper well documented and is the documentation provided alongside the assets?
    \item[] Answer: \answerNA{} 
    \item[] Justification: This paper does not release new assets.

\item {\bf Crowdsourcing and research with human subjects}
    \item[] Question: For crowdsourcing experiments and research with human subjects, does the paper include the full text of instructions given to participants and screenshots, if applicable, as well as details about compensation (if any)? 
    \item[] Answer: \answerNA{} 
    \item[] Justification: This paper does not involve crowdsourcing nor research with human subjects.

\item {\bf Institutional review board (IRB) approvals or equivalent for research with human subjects}
    \item[] Question: Does the paper describe potential risks incurred by study participants, whether such risks were disclosed to the subjects, and whether Institutional Review Board (IRB) approvals (or an equivalent approval/review based on the requirements of your country or institution) were obtained?
    \item[] Answer: \answerNA{} 
    \item[] Justification: This paper does not involve crowdsourcing nor research with human subjects.

\item {\bf Declaration of LLM usage}
    \item[] Question: Does the paper describe the usage of LLMs if it is an important, original, or non-standard component of the core methods in this research? Note that if the LLM is used only for writing, editing, or formatting purposes and does \emph{not} impact the core methodology, scientific rigor, or originality of the research, declaration is not required.
    \item[] Answer: \answerNA{} 
    \item[] Justification: LLMs were used only for writing polishing and did not contribute to the core methodology, experimental design, scientific analysis, or originality of this research.

\end{enumerate}

\end{document}